\documentclass[letterpaper, 10 pt, conference]{ieeeconf}
\IEEEoverridecommandlockouts

\usepackage[utf8]{inputenc}

\usepackage{mathbbol}

\usepackage{esvect}

\usepackage[percent]{overpic}

\usepackage{tabu}

\usepackage{multirow}

\def\mc{\multicolumn}

\usepackage[overload]{empheq}

\usepackage{bm}

\usepackage{xcolor}
\definecolor{dg}{rgb}{0.1, 0.6, 0.2}       
\definecolor{b}{rgb}{0.0, 0.0, 1}          

\usepackage{epsfig}
\usepackage{float}
\usepackage{color}
\usepackage{url}

\usepackage{algorithmic}
\usepackage[ruled,linesnumbered,vlined]{algorithm2e}

\usepackage{amssymb, amsfonts}
\usepackage{amsmath}
\usepackage{mathrsfs}

\usepackage{amsthm}
\usepackage{mathtools}

\let\labelindent\relax
\usepackage{enumitem}       
\setenumerate[enumerate]{align=left}

\usepackage{graphicx}
\usepackage{subcaption}
\usepackage{caption}

\usepackage{cite}

\makeatletter
\let\NAT@parse\undefined
\makeatother
\usepackage{hyperref} 
\usepackage{footmisc}   

\usepackage{balance}

\newcommand{\norm}[1]{\left\lVert#1\right\rVert}
\newcommand{\abs}[1]{\left\lvert#1\right\rvert}

\makeatletter
\newlength\tmp@\newlength\t@mp
\newcommand{\comp}[3]
  {\mathop{ \settowidth\tmp@{$\displaystyle\mathop{#1}^{#3}_{#2}$}
  \hbox to \tmp@{\hss \settowidth\t@mp{$\displaystyle #1$}\setlength\t@mp{.45\t@mp}
  $\displaystyle\mathop{#1}^{\hspace\t@mp #3}_{\hspace{-\t@mp}#2}$
  \hss} }}
\makeatother

\newcommand{\Int}[2]
{\int_{#1}^{#2}}

\DeclareMathOperator*{\argmin}{argmin}

\hypersetup{
    colorlinks=true,
    linkcolor=blue,
    filecolor=blue,      
    urlcolor=blue,
    citecolor=blue,
}

\def\d{\delta}

\def\o{\omega}
\def\s{\sigma}

\def\D{\Delta}

\def\G{\Gamma}

\def\R{\mathbb{R}}
\def\N{\mathbb{N}}

\def\l{\left}
\def\r{\right}

\def\quat{\mathbf{q}}

\def\pos{\mathbf{p}}
\def\vel{\mathbf{v}}

\newcommand{\fr}[1]{\texttt{#1}}

\newcommand{\vbf}[1]{{\bm{\mathbf{#1}}}}

\def\resi{\bm{r}}

\def\bias{\mathbf{b}}

\def\rot{\mathbf{R}}
\def\tf{\mathbf{T}}
\def\trans{\mathbf{t}}
\def\SO{\mathrm{SO(3)}}

\def\Exp{\mathrm{Exp}}
\def\Log{\mathrm{Log}}

\def\Dt{ {\D t} }
\def\dt{ {\d t} }
\def\wrt{\text{w.r.t. }}

\def\X{\mathcal{X}}
\def\Y{\mathcal{Y}}
\def\Z{\mathcal{Z}}
\def\U{\mathcal{U}}
\def\I{\mathcal{I}}
\def\V{\mathcal{V}}
\def\M{\mathcal{M}}
\def\L{\mathcal{L}}

\def\F{\mathcal{F}}
\def\C{\mathcal{C}}
\def\K{\mathcal{K}}
\def\T{\mathcal{T}}

\def\fB{\fr{B}}
\def\fC{\fr{C}}
\def\fL{\fr{L}}
\def\fW{\fr{W}}

\def\Xhat{\hat{\X}}
\def\That{\hat{\T}}

\def\f{\vbf{f}}
\def\n{\vbf{n}}

\def\dis{\vbf{d}}

\newcommand{\ul}[1]{\underline{#1}}
\newcommand{\tb}[1]{\textbf{#1}}

\newcommand{\tref}[1]{Tab. \ref{#1}}
\newcommand{\fref}[1]{Fig. \ref{#1}}

\begin{document}

\title{\bf VIRAL SLAM: Tightly Coupled Camera-IMU-UWB-Lidar SLAM}
\author{
        Thien-Minh Nguyen, \IEEEmembership{Member,~IEEE},
        Shenghai Yuan,
        Muqing Cao,\\
        Thien Hoang Nguyen, \IEEEmembership{Student Member,~IEEE}
		and Lihua Xie, \IEEEmembership{Fellow,~IEEE}
\thanks{This work was supported by the Wallenberg AI, Autonomous Systems and Software Program (WASP) funded by the Knut and Alice Wallenberg Foundation, under the Grant Call 10013 - Wallenberg-NTU Presidential Postdoctoral Fellowship 2020. (Corresponding Author: Thien-Minh Nguyen)}%
\thanks{$^{1}$The authors are with School of Electrical and Electronic Engineering, Nanyang Technological University, Singapore 639798, 50 Nanyang Avenue. (e-mail:
{\tt\footnotesize \{thienminh.nguyen@, shyuan@, mqcao@, e180071@e., elhxie@\}ntu.edu.sg}).}%
}


\maketitle

\thispagestyle{plain}
\pagestyle{plain}

\begin{abstract}
In this paper, we propose a tightly-coupled, multi-modal simultaneous localization and mapping (SLAM) framework, integrating an extensive set of sensors: IMU, cameras, multiple lidars, and Ultra-wideband (UWB) range measurements, hence referred to as VIRAL (visual-inertial-ranging-lidar) SLAM.
To achieve such a comprehensive sensor fusion system, one has to tackle several challenges such as data synchronization, multi-threading programming, bundle adjustment (BA), and conflicting coordinate frames between UWB and the onboard sensors, so as to ensure real-time localization and smooth updates in the state estimates.
To this end, we propose a two stage approach. In the first stage, lidar, camera, and IMU data on a local sliding window are processed in a core odometry thread. From this local graph, new key frames are evaluated for admission to a global map. Visual feature-based loop closure is also performed to supplement the global factor graph with loop constraints. When the global factor graph satisfies a condition on spatial diversity, the BA process will be triggered to update the coordinate transform between UWB and onboard SLAM systems. The system then seamlessly transitions to the second stage where all sensors are tightly integrated in the odometry thread. The capability of our system is demonstrated via several experiments on high-fidelity graphical-physical simulation and public datasets.

\end{abstract}


\IEEEpeerreviewmaketitle

\section{Introduction}

Localization is arguably one of the most important capabilities for mobile robots, especially for Unmanned Aerial Vehicles (UAVs). Obviously, a common approach to ensure reliable and accurate localization is to combine multiple sensors for their complementary advantages as well as redundancy.
For example, since lidar is not affected by lighting condition or lack of visual features, which can easily destabilize most visual-inertial-odometry (VIO) systems, the robot can still rely on this type of sensor for localization in low light or low-texture conditions. In addition, camera can also enable loop closure capability, while lidar pointcloud map can help augment the visual feature's depth estimation process \cite{zhang2018laser, shan2021lvisam}. This is one of the main advantages that motivates us to develop a tightly coupled camera-lidar-IMU-based Simultaneous Localization and Mapping (SLAM) system in this paper.

\begin{figure}
    \centering
    \begin{overpic}[width=\linewidth,
                    ]{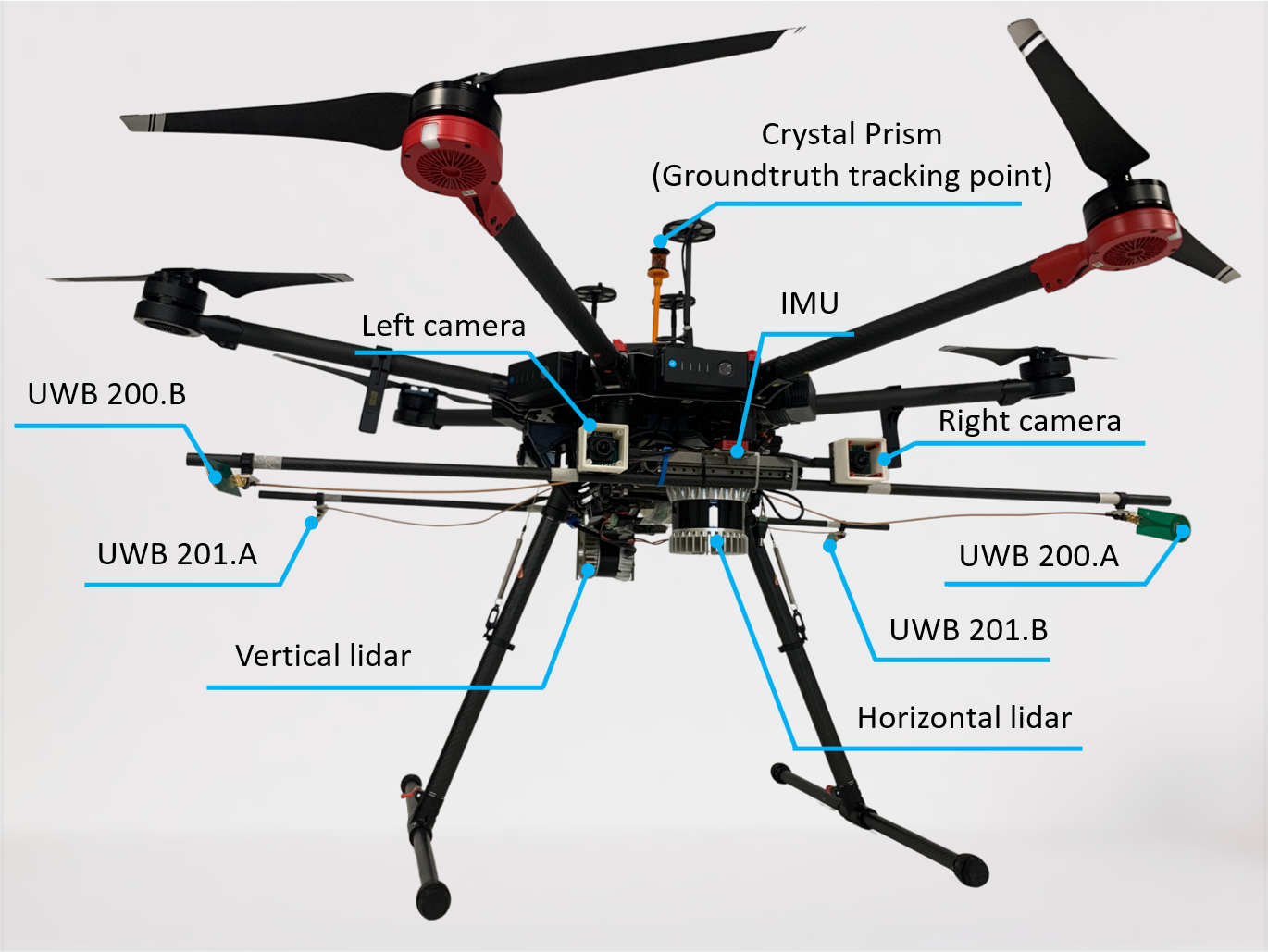}
	\end{overpic}
	\caption{Hardware setup of the VIRAL SLAM system on a UAV: a hardware-synchronized stereo camera rig, a 400 Hz IMU, two 16-channel lidars, four body-offset UWB ranging nodes and a crystal prism that is tracked by a Leica total station for millimeter-accuracy groundtruth. }
	\label{fig: hardware}
\end{figure}

Besides the aforementioned benefits of a visual-lidar localization system, integration of Ultra-wideband (UWB) into the SLAM system can also provide another layer of backup in case both lidar and camera lose track, and also allows user to obtain global localization information relative to the inspected object \cite{nguyen2020liro, nguyen2021viral}. However, to successfully integrate UWB with SLAM, especially in the real-time localization process, one must first estimate the coordinates of the anchors in the SLAM coordinate frame $\fL$, which is the methodology used in previous works \cite{nguyen2020tightlyauro, nguyen2021range}. In this paper, we propose a new approach. Specifically, using the distance measurements between the anchors, we can set up a nominal coordinate of the anchors, which effectively defines a preferred frame $\fW$ that aligns with the mission to be conducted in the environment (more details in Sec. \ref{sec: coord}), and then further refine the transform between $\fL$ and $\fW$ in the BA process. Subsequently, the anchors' coordinates in $\fW$ can be converted to $\fL$ and used for constructing the range-based factors in the optimization process over the local sliding window. The separation of estimating the anchor coordinates and estimating the robot states is a deliberate choice to ensure convergence, especially in the case when the movement in the sliding window is too short, which lacks excitation for convergence of the anchor position estimates. It is also noted that we focus on a simple yet effective UWB network of two or three anchors, with multiple body-offset ranging nodes in the UAV. This simple network allows relatively accurate initialization of the robot and anchor position in $\fW$, which facilitates accurate and seamless integration of UWB into the SLAM system.

The contribution of our work can be stated as follows:
\begin{itemize}
    \item We propose a comprehensive SLAM framework that tightly integrates multiple sensors of different sensing modalities, i.e. lidars, cameras, IMU and UWB ranging sensors, in a seamless manner.
    \item We propose a map-matching marginalization (MMM) scheme for visual features using the local map constructed from lidar pointclouds.
    \item We devise a loop closure scheme that is triggered by visual place recognition and further refined via a two-stage pointcloud alignment.
    \item We propose a novel scheme to fuse UWB, where estimation of the anchor position and ranging bias is delegated to the bundle adjustment (BA) thread, and their values are fixed in the local sliding window optimization for the fusion of UWB range.
    \item We conduct extensive experiments to validate VIRAL SLAM and compare it with other state-of-the-art methods in a variety of real-world scenarios and high-fidelity simulations.
\end{itemize}

\section{Related Works}

To the best of our knowledge, our work features a tightly coupled SLAM system that integrates one of the most comprehensive sensor suites. While localization methods based on mainly camera or lidar (with or without IMU) are abundant, only a handful of works have investigated tightly coupled visual and lidar information in the literature. In \cite{zhang2018laser}, Zhang et al proposed a method where VIO and lidar data were employed in a cascaded manner. In this framework, high rate VIO data is used to help initialize the scan-matching process, while pointcloud map can be used to help retrieve the visual feature's depth. On the other hand, in \cite{zuo2019lic, zou2020lic} Zuo et al proposed an MSCKF framework to asynchronously update the robot states when lidar and visual features are obtained, and IMU is used for propagation in between of these states. In \cite{wisth2021unified}, stereo camera, IMU, lidar were fused together on a unified framework, synchronized with camera data. We note that loop closure and BA are not considered in the aforementioned works (as opposed to our proposed VIRAL SLAM), therefore drift is still an intrinsic problem in these approaches.

To address the drift issue, in \cite{graeter2018limo}, Graeter et al studied the problem of estimating the scale of a pose graph obtained from monocular visual SLAM with lidar information. In \cite{shao2019stereo}, Shao et al considered an approach similar to \cite{zhang2018laser}, but also used camera and ICP for loop closure. However they require hardware-synchronized camera-lidar messages, which can only produce very low rate data. In \cite{shan2021lvisam}, Shan et al proposed a loose integration of VIO output from VINS-Mono \cite{qin2017vins} to LIO-SAM \cite{shan2020liosam}, which itself loosely integrates LeGO-LOAM output \cite{shan2018lego} with IMU preintegration, in a gtsam pose-graph optimization framework. This approach can be unreliable as the whole chain depends on whether the core lidar-based process runs well. If there is a low-texture case when lidar localization is unstable, its error can reverberate up the chain, which appears to be the case in some of our experiments. We also note that the aforementioned works \cite{zhang2018laser, graeter2018limo, shao2019stereo, shan2021lvisam, shan2020liosam, shan2018lego} do not consider the integration of multiple lidars like VIRAL SLAM. On the other hand, while the MLOAM and BLS methods \cite{jiao2020robust, chen2021low} did address this issue, they focused purely on lidar and no camera and IMU is involved. In \cite{nguyen2021miliom}, we proposed a multi-input lidar-inertia odometry and mapping scheme called MILIOM, which clearly demonstrates the robustness, accuracy, and real-time performance. VIRAL SLAM system is developed based on this lidar-based system.
 
Another trend in the literature is the use of UWB to aid VIO or SLAM process. For example, UWB has been integrated with monocular VIO for drift correction \cite{nguyen2020tightlyauro, nguyen2021range}, or can be used as a variable baseline for cameras on different UAVs \cite{karrer2020distributed}. In recent years, VIO, UWB and lidar have also been used in a loosely coupled manner for relative localization \cite{nguyen2019integrated, nguyen2019distance, nguyen2019persistently, queralta2020vio, xu2012distributed, xu2020decentralized}. In our previous work \cite{nguyen2021viral}, a tightly-coupled fusion of body-offset range measurement with IMU preintegration and pose displacement derived from LOAM/VINS subsystems was proposed. We showed that the system can achieve better accuracy compared to traditional onboard self-localization methods. Based on this work, tight coupling of UWB with lidar and IMU preintegration factors was investigated in \cite{nguyen2020liro}. However, this preliminary work is still restrictive in that the lidar processing pipeline inherited from LIO-Mapping \cite{ye2019tightly} was quite inefficient, and no visual information was considered.


The remainder of the paper is organized as follows: in Sec. \ref{sec: preliminary}, we lay out some basic definitions and our general approach towards synchronization. Sec. \ref{sec: localization}
then presents the main function blocks that process the sensor data for the local sliding window optimization, while Sec. \ref{sec: mapping} goes into detail of the global map management, which includes loop closure and BA processes. We demonstrate the capability of our method via several experiments on public and simulated datasets in Sec. \ref{sec: experiment}. Finally, Sec. \ref{sec: conclusion} concludes our work.

\section{Preliminaries} \label{sec: preliminary}



\subsection{Coordinate frames} \label{sec: coord}

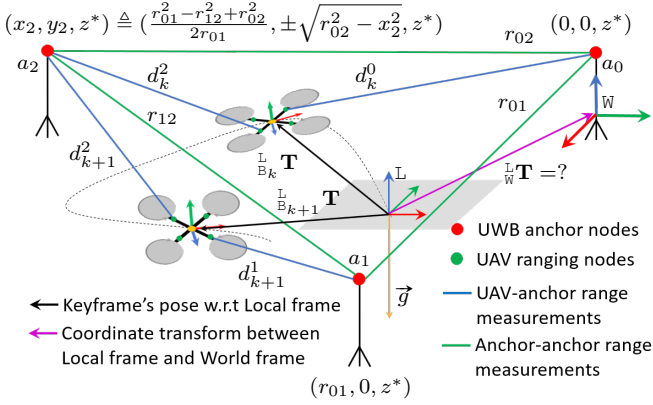
\begin{figure}[h]
    \centering
    \vspace{0.5cm}
    \begin{overpic}[width=\linewidth,
                        ]{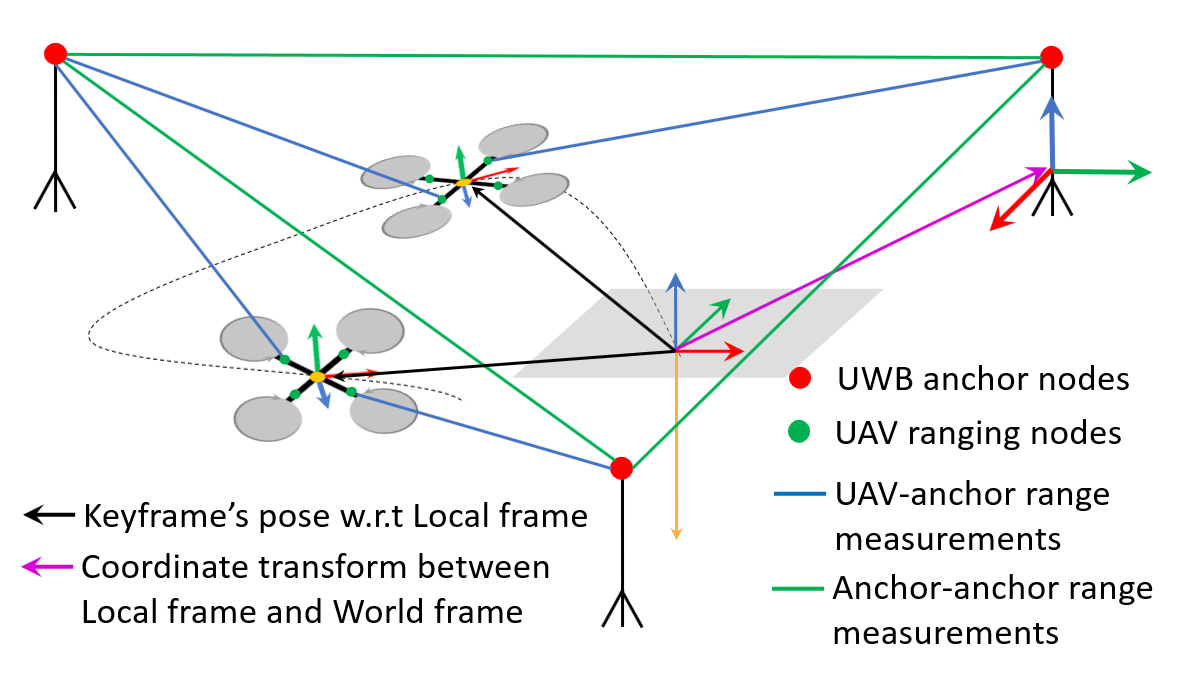}
                        
                        \put(00.00, 50.00){\footnotesize $a_2$}
                        \put(90.00, 50.00){\footnotesize $a_0$}
                        \put(51.00, 20.00){\footnotesize $a_1$}
                        
                        \put(-2.00, 56.00){\footnotesize $(x_2, y_2, z^*) \triangleq (\frac{r_{01}^2 - r_{12}^2 + r_{02}^2}{2r_{01}}, \pm \sqrt{r_{02}^2 - x_2^2}, z^*)$}
                        \put(82.00, 56.00){\footnotesize $(0, 0, z^*)$}
                        \put(45.00,  0.00){\footnotesize $(r_{01}, 0, z^*)$}
                        
                        \put(75.00, 54.50){\footnotesize $r_{02}$}
                        \put(20.00, 42.00){\footnotesize $r_{12}$}
                        \put(74.00, 44.00){\footnotesize $r_{01}$}
                        
                        \put(52.00, 48.00){\footnotesize $d_{k}^0$}
                        \put(20.00, 48.00){\footnotesize $d_{k}^2$}
                        
                        \put(08.00, 36.00){\footnotesize $d_{k+1}^2$}
                        \put(34.00, 17.50){\footnotesize $d_{k+1}^1$}
                        
                        \put(58.00, 33.00){\footnotesize $\fL$}
                        \put(90.00, 44.00){\footnotesize $\fW$}
                        
                        \put(58.00, 14.00){\footnotesize $\vv{g}$}
                        
                        \put(37.00, 34.50){\footnotesize ${}_{\fB_k}^{\fL}\tf$}
                        \put(40.00, 29.00){\footnotesize ${}_{\fB_{k+1}}^{\fL}\tf$}
                        \put(75.00, 33.00){\footnotesize ${}_{\fW}^{\fL}\tf = ?$}
	\end{overpic}
	\caption{A so-called world frame $\fW$ can be defined when fixing the coordinates of the anchor nodes using the anchor-to-anchor distances. On the other hand, the SLAM system takes reference to a local coordinate frame $\fL$ that coincides with the initial key frame's pose. To successfully combine UWB with SLAM, the transform ${}^\fL_\fW\tf$ needs to be resolved.}
	\label{fig: coordinates}
\end{figure}

In this paper, we define a so-called \textit{local frame} $\fL$ whose origin coincides with the position of the body frame at the initial time, and the $z$ axis points to the opposite direction of gravity, and the robot's initial yaw angle is zero.
In addition to $\fL$, we fix the coordinates of the anchors, which defines another so-called \textit{world frame} $\fW$. Since three anchors reside on a plane, the 3D coordinates of these anchors in $\fW$ can be determined via the distances between the anchors, plus a nominal height $z^*$. Fig. \ref{fig: coordinates} describes our coordinate systems in more details.

\subsection{State estimates}

At each time step $t_k$, we define a sliding window $\That_k$ consisting of the robot's state estimates over the last $M$ time steps as follows:
\begin{alignat}{2}
    &\That_k &&= \l(\Xhat_{w}, \Xhat_{w+1}, \dots, \Xhat_{k}\r), w \triangleq k - M + 1. \label{eq: X hat k to k+M}\\
    &\Xhat_k &&= \Big( \hat{\quat}_k, \hat{\pos}_k, \hat{\vel}_k, \hat{\bias}^{\o}_{k}, \hat{\bias}^{a}_{k}\Big) \in \SO \times \R^{12}, \label{eq: X hat k}
\end{alignat}
where $\hat{\quat}_k$, $\hat{\pos}_k$, $\hat{\vel}_k$ are respectively the orientation quaternion, position and velocity state estimates \wrt the local frame $\fL$ at time $t_k$; $\hat{\bias}^{a}_{k}, \hat{\bias}^{\o}_{k}$ are respectively the IMU accelerometer and gyroscope biases.

Besides, we also define the states for the inverse depth of $N_\V^k$ visual features being tracked on the sliding window as:
\begin{equation}
    \hat{\lambda}^1, \hat{\lambda}^2, \dots \hat{\lambda}^{N_\V^k},\ \hat{\Lambda}_k \triangleq (\hat{\lambda}^1, \hat{\lambda}^2, \dots \hat{\lambda}^{N_\V^k}) \in \R^{N_\V^k}
\end{equation}


\subsubsection{Global pose graph and UWB parameters}

A global pose graph is developed with marginalized key frames from the sliding window estimation process. For each key frame $i$ stored in the memory, we define its pose estimate as ${}^{\fL}\hat{\tf}_i$. The pose estimates will be updated in the BA process whenever a certain number of new key frames are admitted, or a loop factor is obtained.
Besides the key frame pose, we also seek to estimate the following UWB-related parameters:
\begin{equation}  \label{eq: uwb extrinsic intrinsic}
    {}^\fL_\fW\tf = ({}^\fL_\fW\rot, {}^\fL_\fW\pos),\ {}^\fL_\fW\rot \in \SO,\ {}^\fL_\fW\pos \in \R^3;\ \bias^r \in \R,
\end{equation}
where ${}^\fL_\fW\tf$ is the coordinate transform between the local and world frames that were introduced in Sec. \ref{sec: coord}, and $\bias^r$ is the ranging bias that is present in our problem due to the use of extension cables to place the UWB ranging nodes at different points on the robot \cite{nguyen2018robust}. Taking VIO as an analogy, ${}^\fL_\fW\tf$ and $\bias^r$ are similar to the extrinsic and intrinsic parameters of the UWB ranging and communication network.

\subsection{Synchronization} \label{sec: sync}

\begin{figure}[h]
    \centering
    \begin{overpic}[width=\linewidth,
                    ]{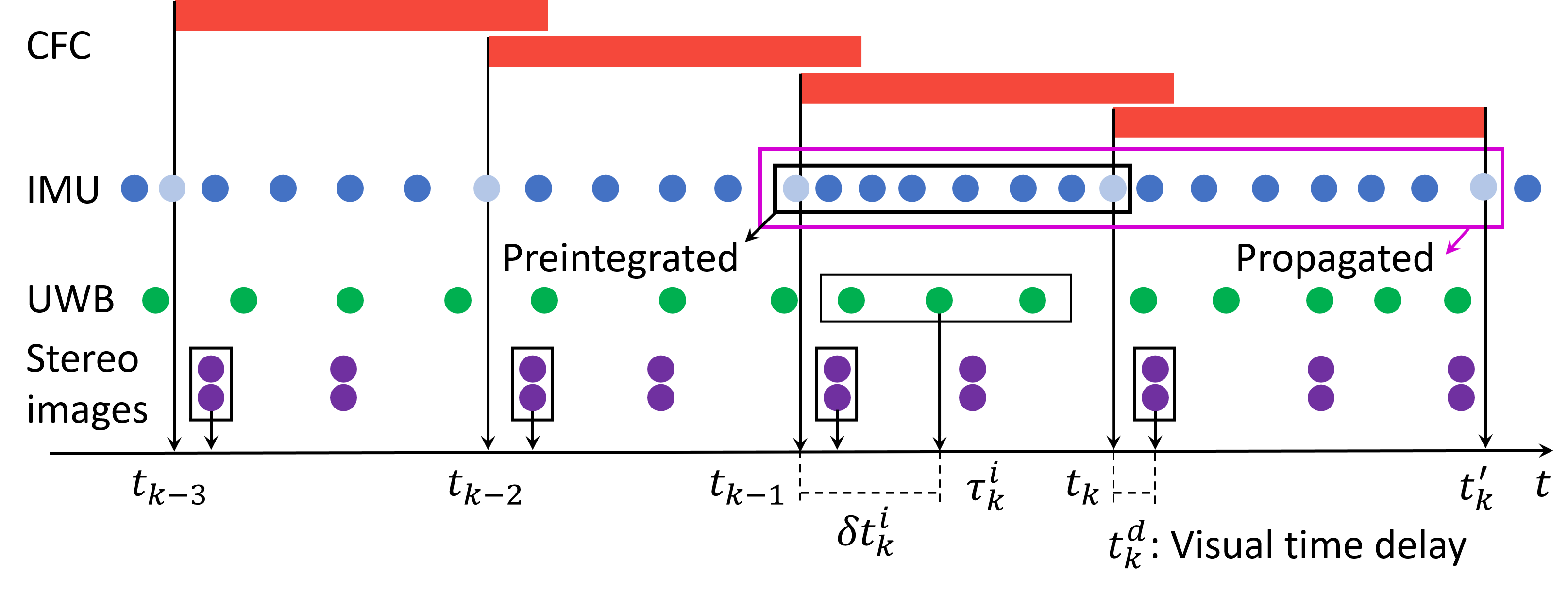}
	\end{overpic}
	\caption{Synchronization among the sensors. The light blue circles represent the interpolated IMU samples.}
	\label{fig: lidar imu img sync}
\end{figure}

In this work, our synchronization scheme is an combination of previous schemes for multiple lidars with IMU \cite{nguyen2021miliom} and lidar with IMU and UWB data
\cite{nguyen2020liro}, with stereo-images being the new addition. Fig. \ref{fig: lidar imu img sync} is an illustration of our synchronization scheme.

Briefly speaking, one lidar is arbitrarily chosen as the primary whose timestamps are used to determine the sliding window's time steps, and other lidars' inputs are merged into the primary lidar's, yielding a combined feature cloud (CFC) as a single sensor input. IMU data are associated with the time steps for propagation and preintegration.
For UWB samples, they are grouped into "bundles" based on the intervals between the time steps. Here we denote the timestamp of a UWB sample as $\tau^i_k$, which implies that $\tau^i_k \in (t_{k-1},\ t_k]$. Knowing this will allow us to associate the sample with the correct state in the construction of the cost factor in the later part.

For the cameras, they are triggered by an external hardware apparatus, thus their images can be synchronized into pairs before further synchronized with the lidar CFCs. For each time step, we admit the image pair that is closest in time to it, and measure the time delay $t^d_k$. This time delay will be used to compensate for the visual feature's pixel coordinate when they are tracked in the image plane.

Fig. \ref{fig: lidar imu img sync} is a snapshot of the sliding window at time $t$, where all of the sensor data needed for constructing the cost function in the local sliding window optimization block are available. After the optimization process elapses, we can obtain optimized states $\hat{\X}_w, \hat{\X}_{w+1}, \dots \hat{\X}_k$ and nominate one of them as a key frame candidate $\K$ to the global map. This is the snapshot of the system as shown in Fig. \ref{fig: overview}.

\subsection{System overview}

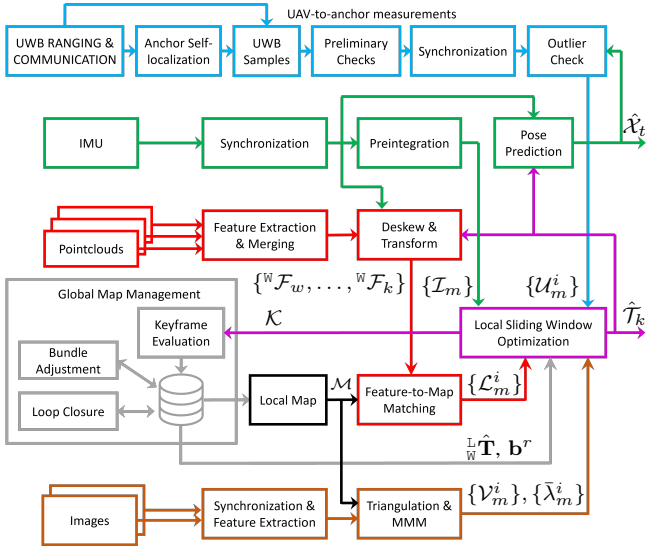
\begin{figure}[h]
    \centering
	\begin{overpic}[width=\linewidth,
                        ]{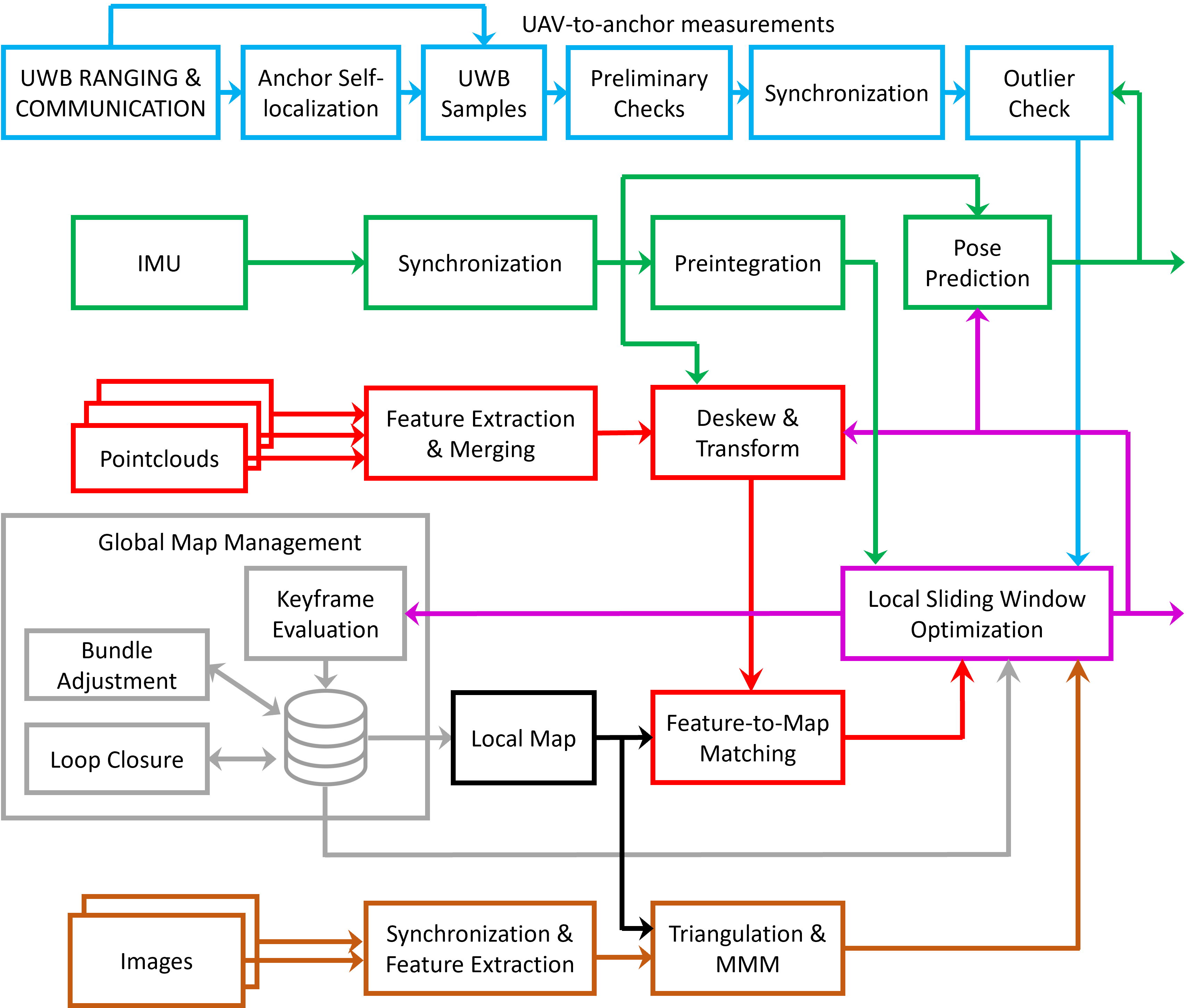}
                        \put(95.0,  63.5){\footnotesize  $\Xhat_{t}$}
                        \put(95.0,  34.5){\footnotesize  $\That_{k}$}
                        \put(50.0,  23.0){\footnotesize  ${}^\M$}
                        \put(40.0,  34.0){\footnotesize  $\K$}
                        \put(38.0,  39.5){\footnotesize  $\{{}^\fW\F_w, \dots,{}^\fW\F_k\}$}
                        \put(64.0,  39.0){\footnotesize  $\{\I_m\}$}
                        \put(80.0,  39.0){\footnotesize  $\{\U^i_m\}$ }
                        \put(71.0,  24.0){\footnotesize  $\{\L^i_m\}$}
                        \put(71.0,  14.0){\footnotesize  ${}^\fL_\fW\hat{\tf}$, $\bias^r$}
                        \put(71.0,  7.0){\footnotesize  $\{\V^i_m\},$}
                        \put(81.0,  7.0){\footnotesize  $\{\bar{\lambda}^i_m\}$}
	\end{overpic}
	\caption{Overview of the VIRAL SLAM system. The colors are used to distinguish those lines that intersect each other but do not connect.}
	\label{fig: overview}
\end{figure}

Fig. \ref{fig: overview} presents the main function blocks of our VIRAL SLAM system. Most expansive of all is the real-time localization thread, where all sensor data are synchronized and processed to eventually create factors in a cost function that is optimized using the ceres solver\cite{ceres-solver}. Besides this time-critical thread, another thread runs in the background to manage the key frames, detect loop closure, and BA optimization. The details of these blocks will be described in the next sections.

\section{Real-Time Localization Function Blocks} \label{sec: localization}

\subsection{Local sliding window optimization}

To estimate the states on the local sliding window, we seek to construct and optimize the following cost function:
\begin{align}
    &f(\That_k, \hat{\Lambda}_k) \triangleq
    \Bigg\{\sum_{m = w+1}^{k} \norm{\resi_\I(\hat{\X}_{m-1}, \hat{\X}_{m}, \I_m)}^2_{\vbf{P}_{\I_m}^{-1}} \nonumber\\
    &+ \sum_{m = w}^{k} \sum_{i = 1}^{N_\L^m}\rho_H\l(\norm{\resi_\L(\hat{\X}_{m}, \L_m^i)}^2_{\vbf{P}_{\L_m^i}^{-1}}\r)
    \nonumber\\
    &+ \sum_{m = w}^{k} \sum_{i = 1}^{N_\U^m}\norm{\resi_{\U}(\hat{\X}_{m-1}, \hat{\X}_{m}, {}^{\fL}_{\fW}\hat{\tf}, \hat{\bias}^r, \U^i_m)}_{\vbf{P}_{\U_m^i}^{-1}}
    \nonumber\\
    &+ \sum_{i = 1}^{N_\V^k}\sum_{b\in\C^i}\rho_A\l(\norm{\resi_\V(\hat{\X}_{m_a}, \hat{\X}_{m_b}, \tilde{\lambda}^i, \V^i_{ab})}^2_{\vbf{P}_{\V^i_{ab}}^{-1}}\r)
    \Bigg\},
    \label{eq: cost function}
\end{align}
where $\rho_H(\cdot)$ and $\rho_A(\cdot)$ are the Huber and arctan loss functions used to reduce the effects of outliers; $\I_m$, $\L^i_m$, $\U^i_m$, $\V^i_{ab}$ are the elementry observations from IMU, Lidar, UWB and visual feature, respectively; $N_\L^m \in \N$ is the number of \textit{feature-map matching} FMM coefficients extracted from the CFC $\F_m$, $N_\U^m \in \N$ is the number of UWB samples obtained in the interval $(t_{m-1}, t_m]$, $N_\V^k \in \N$ is the number of visual features that are tracked on the sliding window from $t_w$ to $t_k$, and $\C^i$ refers to the set of cameras that observe the visual feature $\f^i$, excluding $\fC_a$, $\tilde{\lambda}^i$ can be either the state estimate $\hat{\lambda}^i$ or the marginalized inverse depth $\bar{\lambda}^i$ of the MMM features.
The cost function \eqref{eq: cost function} summarizes the coupling of each sensor's factor with the state estimate. We will elaborate on how to construct these factors in the next sections.

\subsection{Sensor data processing}

\subsubsection{Lidar \& IMU} \label{sec: lidar processing}
We refer to our previous work \cite{nguyen2021miliom} for the details on how to construct the lidar and IMU factors from the sensor data.

\subsubsection{UWB}
Similar to \cite{nguyen2020liro}, we define each UWB sample as $\U^i_m = \l(\breve{d}^i, {}^\fW\vbf{x}^i, \vbf{y}^i, \tau^i_m, t_{m-1}, t_m\r)$, where $\breve{d}^i$ is the range measurement, ${}^\fW\vbf{x}^i$ is the coordinate of the anchor \wrt $\fW$, $\vbf{y}^i$ is the UAV ranging node in the body frame $\fB_{\tau^i_k}$, $\tau^i_m$ is the message's timestamp, $t_{m-1}$ and $t_m$ are the preceding and succeeding time steps of $\tau^i_m$. However, what is different now is that the distance measurement $\breve{d}^i$ at time $t_k+\dt^i$ is defined by the norm of the vector ${}^{\fL}\dis^i$, corrupted by Gaussian noise and bias as follows:
\begin{align}
    \breve{d}^i &= \norm{{}^\fL\dis^i} + \vbf{\eta}_{\U^i} + \bias^r,\ \vbf{\eta}_{\U^i} \sim \  \mathcal{N}(0, \s_{\U}^2);\\
    {}^{\fL}\dis^i
    &\triangleq \pos_{m} + \rot_{m-1}\Exp\l(s^i\Log(\rot_{m-1}^{-1}\rot_{m})\r) \vbf{y}^i \nonumber\\
    &\qquad\qquad - a^i\vel_{m-1} - b^i{\vel}_{m} - {}^{\fL}_{\fW}\rot\vbf{x}^i - {}^{\fL}_{\fW}\trans, \label{eq: ant-anc displacement}
\end{align}
where $s^i \triangleq \frac{\dt_i}{\Dt_m}$, $a^i \triangleq \frac{\Dt_m^2 - \dt_i^2}{2\D t_m}$, $b^i \triangleq \frac{(\Dt_m - \dt_i)^2}{2\Dt_m}$, $\dt^i \triangleq \tau^i_k - t_{m-1}$, $\Dt_m \triangleq t_m - t_{m-1}$.

Thus, the UWB range residual is defined as:
\begin{align}
    \resi_{\U} \triangleq \| {}^\fL\dis(\hat{\X}_{m-1},\ \hat{\X}_{m},\ {}^{\fL}_{\fW}\hat{\tf},\ \U^i_m)\| + \hat{\bias}^r - \breve{d}^i. \label{eq: uwb residual}
\end{align}


Note that ${}^{\fL}_{\fW}\hat{\tf}$ and $\hat{\bias}^r$ are kept fixed during the sliding window optimization process. Moreover, UWB is not fused until the BA process has updated ${}^{\fL}_{\fW}\hat{\tf}$ and $\hat{\bias}^r$, which is discussed in Sec. \ref{sec: BA}.

\subsubsection{Camera} \label{sec: camera}

Over the sliding window, BRIEF features are tracked and associated with the time steps. For a visual feature $\f^i$ and its pair of projected coordinate $\V^i_{ab} \triangleq \l({}^{\fC_a}\Z^i,\ {}^{\fC_b}\Z^i\r)$ in two cameras $\fC_a$ and $\fC_b$, the residual of this observation is defined as:

\begin{align}\label{eq: visual residual}
    &\resi_\V(\hat{\X}_a, \hat{\X}_b, \V^i_{ab})
    = \pi\l({}^{\fC_b}\hat{\f}^i\r) - {}^{\fC_b}\Z^i,
    \nonumber\\
    &{}^{\fC_b}\hat{\f}^i \triangleq {}^{\fB}_{\fC_b}\rot^{-1}\l(\hat{\rot}_{m_b}^{-1}\l({}^{\fL}\hat{\f}^i - \hat{\pos}_{m_b}\r) - {}^{\fB}_{\fC_b}\trans\r),
    \nonumber\\
    &{}^{\fL}\hat{\f}^i \triangleq \hat{\rot}_{m_a}\l({}^{\fB}_{\fC_a}\rot\l[(\hat{\lambda}^i)^{-1}{}^{\fC_a}\Z^i\r] + {}^{\fB}_{\fC_a}\trans\r) + \hat{\pos}_{m_a},
\end{align}
Note that in this formulation, ${}^{\fB}_{\fC_a}\rot$, ${}^{\fB}_{\fC_b}\rot$, ${}^{\fB}_{\fC_a}\trans$, ${}^{\fB}_{\fC_b}\trans$ are constant extrinsic parameters of the cameras, and the frames $\fC_a$ and $\fC_b$ could be coupled with the same state but different cameras, or different states of the same camera, or both states and cameras are different.

\begin{figure}
    \centering
    \begin{overpic}[width=\linewidth]{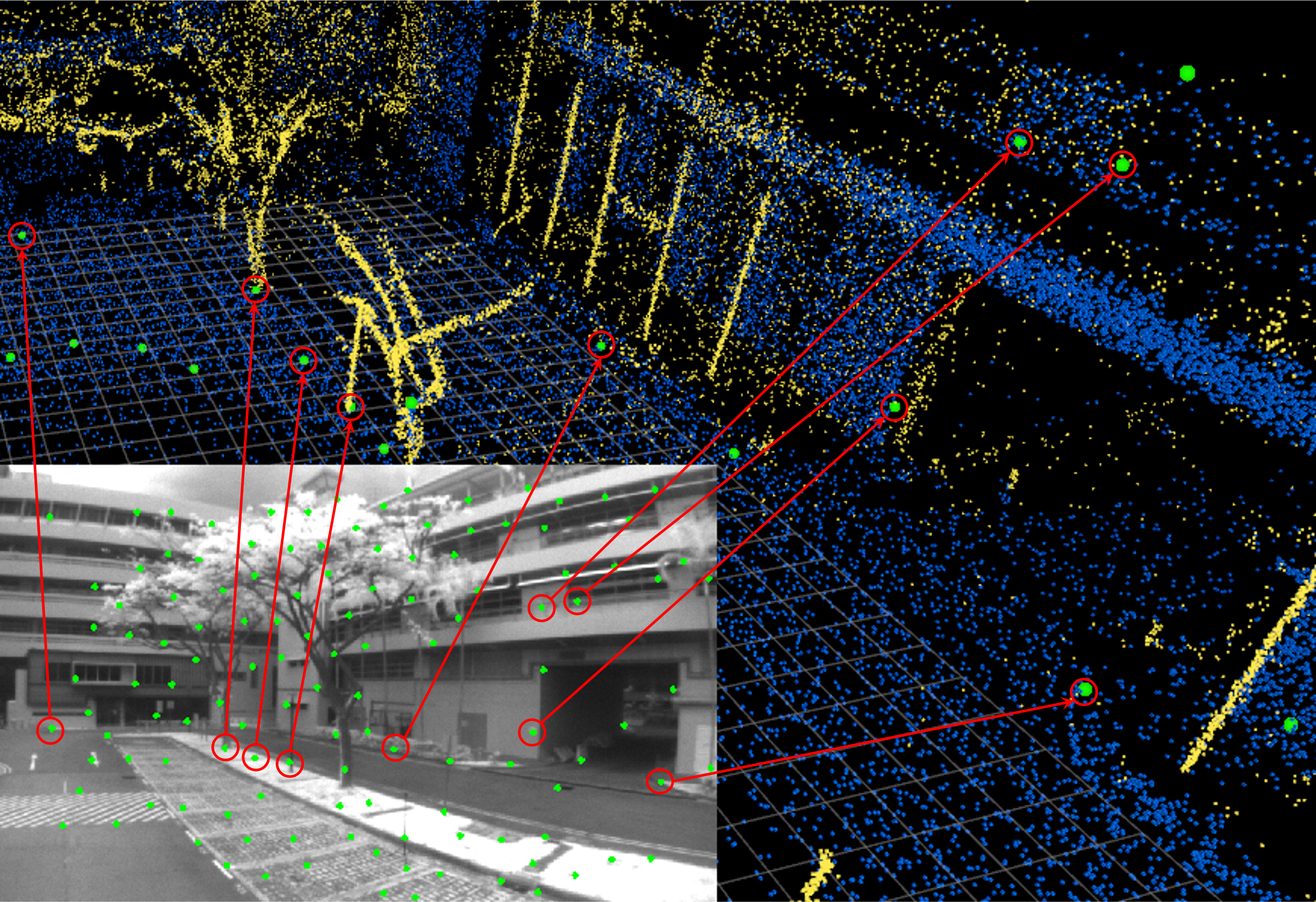}
	\end{overpic}
	\caption{Matching of some visual features from the image to the 3D local map thanks to the MMM process.}
	\label{fig: feature marginalization}
\end{figure}

By checking if a visual feature's estimated 3D coordinates fit well on the 3D local map $\M$, we can marginalize this feature to be a fixed prior in the sliding window. Algorithm \ref{algo: MMM} presents the details of this so-called MMM scheme. Fig. \ref{fig: feature marginalization} illustrates the result of the MMM process for some features.

When a feature is marginalized, its cost factor is similar to \eqref{eq: visual residual}, only that the state estimate of the inverse depth $\lambda^i$ is calculated directly from the marginalized 3D coordinates ${}^\fL\bar{\f}^i$ and kept fixed during the optimization process.

\SetArgSty{textnormal}
\SetKw{Or}{\hspace{\algoskipindent}\itshape {\upshape or}\\}
\SetKw{And}{{\upshape and}}
\SetKwBlock{Condition}{}{}

\begin{algorithm}
    \SetAlgoLined
    \KwIn{${}^{\fC_a}\Z^i$, $\hat{\lambda}^i$, $\hat{\tf}_{m_a}$, $\M$.}
    \KwOut{${}^\fL\bar{\f}^i$.}
    Compute: ${}^{\fC_a}\hat{\f}^i = {}^\fB_{\fC_a}\rot\l[(\hat{\lambda}^i)^{-1}{}^{\fC_a}\Z^i\r] + {}^\fB_{\fC_a}\trans$\;
    \label{algo: MMM feature's coordinates in cam}
    Compute: ${}^{\fL}\hat{\f}^i = \hat{\rot}_{m_a}{}^{\fC_a}\hat{\f}^i + \hat{\pos}_{m_a}$\;
    \label{algo: MMM feature's coordinates in world}
    Find $\mathcal{N} = \text{KNN}({}^{\fL}\hat{\f}^i,\ \M)$\; \label{algo: MMM KNN}
        Find $\n^* = \argmin_{\n \in \R^3} \sum_{\vbf{x} \in \mathcal{N}} ||\n^\top \vbf{x} + 1||^2$\;
        \label{algo: MMM plane}
            Compute: $\bar{\n} = \frac{\n^*}{\norm{\n^*}}$ and ${}^\fL_{\fC_a}\hat{\pos} = \hat{\rot}_{m_a}{}^\fB_{\fC_a}\trans + \hat{\pos}_{m_a}$\; \label{algo: MMM cam pos}
            Compute: $\vbf{u} = \hat{\rot}_{m_a}{}^\fB_{\fC_a}\rot{}^{\fC_a}\Z^i$\; \label{algo: MMM project dir}
            Compute: $\vbf{c} = {}^\fL_{\fC_a}\hat{\pos} - \vbf{u}(1 + \bar{\n}^{\top} {}^\fL_{\fC_a}\hat{\pos})/(\bar{\n}^{\top}\vbf{u})$\; \label{algo: MMM intersect}
        \If{ $\min_{\vbf{x} \in \mathcal{N}} ||\vbf{x} - {}^{\fL}\hat{\f}^i|| \leq 0.25$ \label{algo: MMM check nearest nb}
             \Condition{\And $\norm{\vbf{c} - \vbf{x}} < 1.0,\ \forall \vbf{x} \in \{{}^{\fL}\hat{\f}^i\} \cup \mathcal{N}$ \label{algo: MMM close nbb}\\
                        \And $\abs{\bar{\n}^\top \vbf{x} + 1} < 0.1,\ \forall \vbf{x} \in \mathcal{N}$} \label{algo: MMM plane well fit}
           }
        {
            Set: ${}^\fL\bar{\f}^i = \vbf{c}$\;
        }
  \caption{MMM process on a visual feature ${}^{\fC_a}\f^i$}
  \label{algo: MMM}
\end{algorithm}

\section{Global Optimization Blocks}\label{sec: mapping}

\subsection{Key frame management}

\subsubsection{Key frame admission}
The key frame admission procedure is similar to our previous work \cite{nguyen2021miliom}. Briefly speaking, after each optimization on the sliding window, we find a number of nearest neighbors of the state at time $t_{k-M/2}$ and if the relative distance or relative rotation to all of these neighbours exceed a certain threshold, the information associated with this time step will be marginalized as prior.

\subsubsection{Key frame selection}

The selection of the key frames is needed for construction of a local pointcloud map for FMM process. This selection process takes place before the optimization process and is based on the IMU-propagated pose $\breve{\tf}_k$. Hence, the set of these key frames is a union of $ \{\K_a\} \cup \{\K_b\} \cup \{\K_c\}$, where $\{\K_a\}$ is the set of the last $M$ key frames, $\{\K_b\}$  is the set of $M$ nearest neighbors of $\breve{\tf}_k$, and $\{\K_c\}$ is the set of key frames representing their $2m \times 2m \times 2m$ voxel cells that are within a radius from $\breve{\tf}_k$.

\subsection{Bundle Adjustment} \label{sec: BA}

For the BA process, our task is to construct and optimize the following cost function:
\begin{align}
    &f(\hat{\Y}, {}^{\fL}_{\fW}\hat{\tf}, \hat{\bias}^r)
    \triangleq
    \Bigg\{\sum_{n = 1}^{N} \norm{\resi_1(\hat{\tf}_{n-1}, \hat{\tf}_{n}, {}^{n-1}_{n}\bar{\tf})}^2_{\vbf{P}_1^{-1}}
    \nonumber\\
    &\qquad+ \sum_{(p, c)\in\mathcal{\mathcal{H}}}\norm{\resi_2(\hat{\tf}_{p}, \hat{\tf}_{c}, {}^p_c\bar{\tf})}^2_{\vbf{P}_{2}^{-1}}
    \nonumber\\
    &\qquad+ \sum_{n = 1}^{N} \sum_{j = 1}^{N_\U^n}\norm{\resi_{3}(\hat{\tf}_{n}, {}^{\fL}_{\fW}\hat{\tf}, \hat{\bias}^r, \bar{\U}^i_n)}_{\vbf{P}_3^{-1}}
    \Bigg\}, \label{eq: ba cost function}
\end{align}
where $\hat{\Y} \triangleq (\hat{\tf}_0, \hat{\tf}_1, \dots \hat{\tf}_N)$ is the key frames' poses, ${}^{n-1}_{n}\bar{\tf}$ and ${}^p_c\bar{\tf}$ are respectively the relative pose and loop closure priors, $\mathcal{H}$ is the set of loop closure pairs, and $\bar{\U}^i_n$ is a marginalized UWB measurement whose timestamp is within 0.2s of the key frame at time $t_n$, i.e.:
\begin{equation}
    \bar{\U}^i_n = \l(\breve{d}^i, {}^\fW\vbf{x}^i, \vbf{y}^i, {}^{t_n}_{\tau^i_n}\bar{\rot}, {}^{t_n}_{\tau^i_n}\bar{\trans}\r),
\end{equation}
where $\breve{d}^i$, ${}^\fW\vbf{x}^i$ $\vbf{y}^i$ are defined similarly to \eqref{eq: ant-anc displacement}, and $({}^{t_n}_{\tau^i_n}\bar{\rot}, {}^{t_n}_{\tau^i_n}\bar{\trans})$ is the relative transform between $\fB_{t_n}$ and $\fB_{\tau^i_n}$, which can be obtained from IMU propagation.

The residuals $\resi_1(\cdot)$, $\resi_2(\cdot)$ over the relative poses are straightforward, while the residual $\resi_{3}$ can be stated as:
\begin{align}
    &\resi_{3}(\hat{\tf}_{n}, {}^{\fL}_{\fW}\hat{\tf}, \hat{\bias}^r, \bar{\U}^i_n)
    = \norm{\hat{\vbf{d}}^i} + \hat{\bias}^r - \breve{d}^i,
    \nonumber\\
    &\hat{\vbf{d}}^i \triangleq \hat{\pos}_n + \hat{\rot}_n{}^{t_n}_{\tau^i_n}\bar{\rot}(\vbf{y}^i + {}^{t_n}_{\tau^i_n}\bar{\trans}) - {}^{\fL}_{\fW}\hat{\rot}\vbf{x}^i - {}^{\fL}_{\fW}\hat{\trans}.
\end{align}

To ensure the pose graph has enough excitation to help the the anchor-related states converge, we do not add the factors of $\resi_{3}$ into the BA cost function right from the beginning. Rather, they are only added in after the spatial distribution of the key frame poses has satisfied a certain condition. Specifically, we calculate the geometric dilution of the key frame positions via the quantity
\begin{equation}
    \G = \l(\sum_{n=1}^N(\bar{\pos}_n - \mu)(\bar{\pos}_n - \mu)^\top\r)^{-1}, \mu = \frac{1}{N}\sum_{n=1}^N\bar{\pos}_n.\ 
\end{equation}
Hence we perform singular value decomposition on $\G$ to obtain the singular values $\s_1 \geq \s_2 \geq \s_3 > 0$. If $\s_1 < c_1$ and $\s_1/\s_3 < c_2$, where $c_1$ and $c_2 > 1$ are some user-defined parameters, then we can start adding the factors $\resi_3(\cdot)$ to \eqref{eq: ba cost function}. Afterwards, we can obtain the estimates of ${}^{\fL}_{\fW}\tf$ and $\bias^r$ which can be used for the fusion of UWB factors in \eqref{eq: cost function}.

\subsection{Loop Closure}

To construct the loop priors ${}^p_c\bar{\tf}$ in Sec. \ref{sec: BA}, a three-stage process is conducted as follows:

\begin{table*}
\centering
\caption{ATE of the SLAM methods over the datasets. The best odometry result is highlighted in \tb{bold}, and the second best is \ul{underlined}. The first nine datasets belong to the NTU VIRAL dataset, the next three belong to the building inspection trials, and the last five are generated from AirSim. The symbol '-' indicates that the method diverges during the experiment.} \label{tab: all ATE}
\renewcommand{\arraystretch}{1.1}
\begin{tabular}{ccccccccc}
\hline\hline
\tb{Dataset}
&\tb{\begin{tabular}[c]{@{}c@{}}VINS-Mono\\ (right camera,\\ odom/BA)\end{tabular}}
&\tb{\begin{tabular}[c]{@{}c@{}}VINS-Fusion\\ (both cameras,\\ odom/BA)\end{tabular}}
&\tb{\begin{tabular}[c]{@{}c@{}}A-LOAM\\ (horz. / vert. / latr.)\end{tabular}}
&\tb{\begin{tabular}[c]{@{}c@{}}LIO-\\SAM\\ (horz.)\end{tabular}}
&\tb{\begin{tabular}[c]{@{}c@{}}MLOAM\\ (all lidars)\end{tabular}}
&\tb{\begin{tabular}[c]{@{}c@{}}VIRAL-SLAM\\ (horz. lidar, \\ odom/BA)\end{tabular}}
&\tb{\begin{tabular}[c]{@{}c@{}}VIRAL-SLAM\\ (all lidars, \\ odom/BA)\end{tabular}}\\ \hline
{eee\_01}
        &{1.650} / {0.568}			
        &{0.608} / {0.306}			
        &{0.212} / {6.827}			
        &{0.075} & {0.249}			
        &\ul{0.064} / {0.084}       
        &\tb{0.060} / {0.086} \\	
{eee\_02}
        &{0.722} / {0.443}			
        &{0.506} / {0.266}			
        &{0.199} / {1.845}			
        &{0.069} & {0.166}			
        &\tb{0.051} / {0.056}       
        &\ul{0.058} / {0.050} \\    
{eee\_03}
        &{1.037} / {0.886}			
        &{0.494} / {0.383}			
        &{0.148} / {3.852}			
        &{0.101} & {0.232}			
        &\ul{0.060} / {0.073}       
        &\tb{0.037} / {0.049} \\	
{nya\_01}
        &{1.475} / {0.830}			
        &{0.397} / {0.237}			
        &{0.077} / {3.206}			
        &{0.076} & {0.123}			
        &\ul{0.063} / {0.061}       
        &\tb{0.051} / {0.058} \\	
{nya\_02}
        &{0.580} / {0.422}			
        &{0.424} / {0.297}			
        &{0.091} / {0.377}			
        &{0.090} & {0.191}			
        &\tb{0.042} / {0.051}       
        &\ul{0.043} / {0.055} \\    
{nya\_03}
        &{1.333} / {0.501}			
        &{0.787} / {0.368}			
        &{0.080} / {0.715}			
        &{0.137} & {0.226}			
        &\ul{0.039} / {0.063}       
        &\tb{0.032} / {0.062} \\    
{sbs\_01}
        &{4.142} / {3.739}			
        &{0.508} / {0.372}			
        &{0.203} / {6.762}			
        &{0.089} & {0.173}			
        &\ul{0.051} / {0.055}       
        &\tb{0.048} / {0.059} \\    
{sbs\_02}
        &{1.605} / {0.890}			
        &{0.564} / {0.369}			
        &{0.091} / {2.496}			
        &{0.083} & {0.147}			
        &\tb{0.056} / {0.062}       
        &\ul{0.062} / {0.052} \\    
{sbs\_03}
        &{1.306} / {0.802}			
        &{0.878} / {0.276}			
        &{0.363} / {3.996}			
        &{0.140} & {0.153}			
        &\ul{0.060} / {0.075}       
        &\tb{0.054} / {0.072} \\    
\hline
{bid\_01}
        &{3.749} / {3.632}			
        &{2.416} / {2.045}			
        &{0.936} / {14.670}			
        &   -    & {4.264}			
        &\ul{0.178} / {0.159}       
        &\tb{0.161} / {0.158} \\    
{bid\_02}
        &{1.257} / {1.238}			
        &{0.837} / {0.603}			
        &{4.359} / {5.043}			
        &   -    & {0.257}			
        &\ul{0.752} / 1.320         
        &\tb{0.343} / {0.603} \\    
{bid\_03}
        &{0.670} / {0.659}			
        &{0.914} / {0.814}			
        &{1.961} / {4.789}			
        &   -    & {3.330}			
        &\ul{2.181} / {1.813}       
        &\tb{0.128} / {0.177} \\    
\hline
{nbh\_01}
        &{4.709} / {4.474}				
        &{1.413} / {1.388}				
        &{86.399} / {53.680} / {53.757} 
        &   -    & {0.321}				
        &\ul{0.149} / {0.200}           
        &\tb{0.146} / {0.194} \\        
{nbh\_02}
        &{3.526} / {2.960}				
        &{2.268} / {1.436}				
        &{47.326} / {40.881} / {40.638} 
        &   -    & {0.369}				
        &\tb{0.084} / {0.142}           
        &\ul{0.096} / {0.162} \\        
{nbh\_03}
        &{3.560} / {2.759}				
        &{1.837} / {0.643}				
        &{6.764} / {50.710} / {50.578} 	
        &   -    & {0.282}				
        &\tb{0.091} / {0.114}           
        &\ul{0.098} / {0.141} \\        
{nbh\_04}
        &{2.707} / {1.981}				
        &{1.974} / {1.513}				
        &{24.448} / {35.747} / {35.970} 
        &   -    & {0.375}				
        &\ul{0.113} / {0.239}           
        &\tb{0.099} / {0.196} \\        
{nbh\_05}
        &{118.644} / {120.532}          
        &{1.255} / {0.825}				
        &{0.834} / {24.624} / {24.852} 	
        &   -    & {0.377}				
        &\ul{0.116} / {0.291}           
        &\tb{0.110} / {0.276} \\        
        \hline\hline
\end{tabular}
\end{table*}

First, when a new key frame is admitted, we compare its visual features with the database using the DBoW library. If a match is flagged, we can extract the transforms $\tf_c$ and $\tf_p$,  referred to as the current and previous key poses, respectively. Then, we search for a number of key frames that were admitted before and after $\tf_p$ to build a local map ${}^{\fB_p}\M_p$ using their corresponding marginalized CFCs, and proceed to the second stage.

At the second stage, we will use ICP to align the CFC ${}^{\fB_c}\F_c$ with ${}^{\fB_p}\M_p$ to obtain a fitness score, as well as an initial guess of ${}^{\fB_p}_{\fB_c}\hat{\tf}$. If the fitness score is below a threshold, we proceed to the third stage.

At the third stage, we perform FMM between ${}^{\fB_c}\F_c$ and ${}^{\fB_p}\M_p$ to calculate the FMM coefficients, then construct the following cost function and optimize it:
\begin{equation} \label{eq: loop cost function}
    f\l({}^{\fB_p}_{\fB_c}\hat{\tf}\r)
    = \sum_{i = 1}^{N_\L^c}\rho\l(\norm{\resi_\L({}^{\fB_p}_{\fB_c}\hat{\tf}, \L_c^i)}^2_{\vbf{P}_{\L_c^i}^{-1}}\r).
\end{equation}
After optimizing \eqref{eq: loop cost function} and obtaining the optimal relative pose ${}^{\fB_p}_{\fB_c}\hat{\tf}^*$, if the ratio $f\l({}^{\fB_p}_{\fB_c}\hat{\tf}^*\r)/N_\L^c$ is below a threshold, ${}^{\fB_p}_{\fB_c}\hat{\tf}^*$ will be registered as a loop closure prior ${}^{\fB_p}_{\fB_c}\bar{\tf}$.

\section{Experiment} \label{sec: experiment}

\subsection{Datasets}
We first employ our recently published NTU VIRAL dataset\footnote{\label{foot: ntuviral website}\url{https://ntu-aris.github.io/ntu\_viral\_dataset/}} \cite{nguyen2021ntuviral}, which features all sensor types covered by VIRAL SLAM.
To further demonstrate the robustness of VIRAL SLAM in low-texture condition,
we conduct further experiments on some building inspections datasets with significant challenges collected near a building facade.
Finally, since no ground truth on the anchor position and the ranging bias are available, to clearly verify this capability of VIRAL-SLAM, we employ AirSim simulator to construct a dataset with absolute ground truth for more accurate evaluate.

\subsection{Comparison}

\begin{figure}[h]
    \centering
    \includegraphics[width=\linewidth]{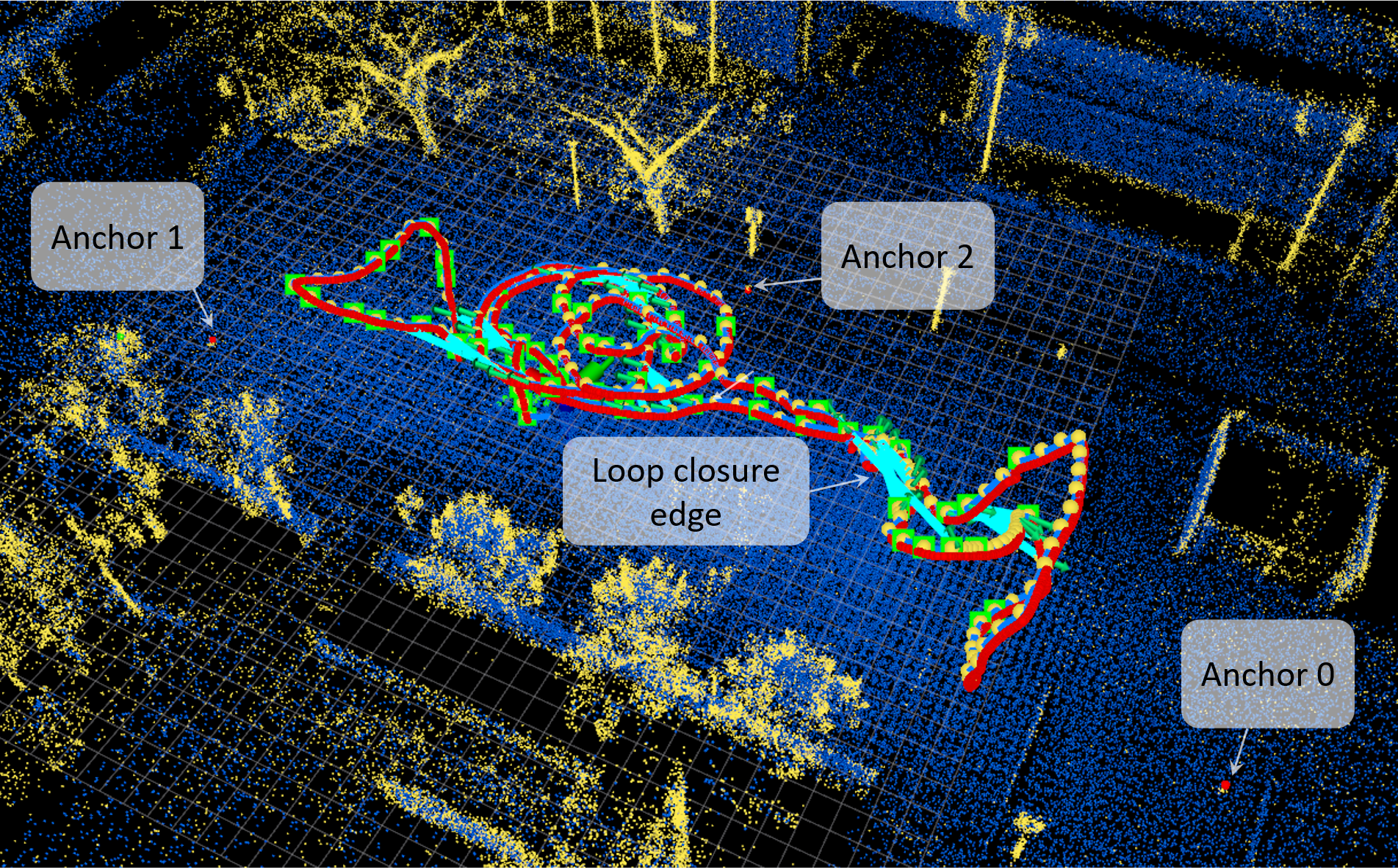}
    \caption{VIRAL SLAM result on eee\_02 dataset. The estimated trajectory is in blue, and ground truth is in red. The key frame poses are marked by the yellow circles. The activated key frames for local map building are highlighted by the green squares. The UWB anchors are also marked with red dots. The loop edges are marked with light cyan lines. Some visual features can be are marked with small green circles. Due to page constraint we refer the readers to the online video recording at \url{https://youtu.be/LerAfvZMb7M} for more detailed illustration of our experiments.}
    \label{fig: viral slam eee 02}
\end{figure}

For comparison VIRAL SLAM, we run other state-of-the-art localization techniques with all of the aforementioned datasets.
All algorithms are run on an NUC 10 computer with core i7 processor.
Each method is slightly modified and configured for their best performance with the dataset. The details of these modified packages can be found on the NTU VIRAL dataset website\footref{foot: ntuviral website}.
Since several lidar-based methods are not designed to for multiple lidars, we also include experiments of VIRAL SLAM using only the horizontal lidar for a fairer comparison. \tref{tab: all ATE} summarizes the Absolute Trajectory Error (ATE) of these methods.

\begin{table}
\centering
\caption{ATE of VIRAL SLAM's key frame positions with different sensor combinations over the employed datasets (IMU is always used). All values are in m. The \textit{average} ATE is calculated for each class of datasets.}
\label{tab: all ATE ablation}
\renewcommand{\arraystretch}{1.1}
\begin{tabular}{ccccc}
\hline\hline
\tb{Dataset}
&\begin{tabular}[c]{@{}c@{}}\tb{Lidars}\end{tabular}
&\begin{tabular}[c]{@{}c@{}}\tb{Lidars}\\ \tb{+Cameras}\end{tabular}
&\begin{tabular}[c]{@{}c@{}}\tb{Lidars}\\ \tb{+UWB}\end{tabular}
&\begin{tabular}[c]{@{}c@{}}\tb{Lidars}\\ \tb{+UWB} \\ \tb{+Cameras}\end{tabular}
\\ \hline

{eee\_01}
        &\tb{0.0380}		
        &\ul{0.0390}		
        &{0.0822}		    
        &{0.0861}\\		    
{eee\_02}
        &\ul{0.0451}		
        &\tb{0.0347}		
        &{0.0647}		    
        &{0.0505}\\		    
{eee\_03}
        &\tb{0.0385}		
        &\ul{0.0438}		
        &{0.0608}		    
        &{0.0494}\\		    
{nya\_01}
        &\tb{0.0429}		
        &\ul{0.0436}		
        &{0.0545}		    
        &{0.0584}\\		    
{nya\_02}
        &\ul{0.0463}		
        &\tb{0.0416}		
        &{0.0635}		    
        &{0.0551}\\		    
{nya\_03}
        &\tb{0.0383}		
        &\ul{0.0392}		
        &{0.0696}		    
        &{0.0621}\\		    
{sbs\_01}
        &\tb{0.0441}		
        &\ul{0.0483}		
        &{0.0585}		    
        &{0.0587}\\	    	
{sbs\_02}
        &{0.0512}	    	
        &\tb{0.0476}		
        &{0.0547}	    	
        &{0.0518}\\	    	
{sbs\_03}
        &\tb{0.0514}		
        &\ul{0.0524}		
        &{0.0696}		    
        &{0.0716}\\		    
{Average}
        &\ul{0.0440}		
        &\tb{0.0434}		
        &{0.0642}		    
        &{0.0604}\\		    
\hline

{bid\_01}
        &{0.2104}		    
        &{0.2039}		    
        &\ul{0.1585}		
        &\tb{0.1583}\\		
{bid\_02}
        &{0.6046}		    
        &\ul{0.6033}		
        &{0.6111}		    
        &\tb{0.6029}\\		
{bid\_03}
        &{0.1904}		    
        &{0.1865}		    
        &\ul{0.1822}		
        &\tb{0.1765}\\		
{Average}
        &{0.3351}		    
        &{0.3312}		    
        &\tb{0.3173}		
        &\tb{0.3126}\\		
\hline
{nbh\_01}
        &\tb{0.0393}		
        &\ul{0.0508}		
        &{0.1910}		    
        &{0.1938}\\		    
{nbh\_02}
        &\ul{0.0374}		
        &\tb{0.0364}		
        &{0.1489}		    
        &{0.1618}\\		    
{nbh\_03}
        &\ul{0.0453}		
        &\tb{0.0342}		
        &{0.1653}		    
        &{0.1413}\\		    
{nbh\_04}
        &\tb{0.0178}		
        &\ul{0.0201}		
        &{0.1811}		    
        &{0.1963}\\		    
{nbh\_05}
        &\tb{0.0225}		
        &\ul{0.0257}		
        &{0.3224}		    
        &{0.2763}\\		    
{Average}
        &\tb{0.0325}		
        &\ul{0.0335}		
        &{0.2017}		    
        &{0.1939}\\		    
\hline\hline
\end{tabular}
\end{table}

\begin{table}
\centering
\caption{Anchor coordinates and the final values estimated by the BA process using the AirSim-generated dataset. All values are in m. Note that the coordinates of anchor 0 is fixed at $(0, 0, 1)$.} \label{tab: anchor estimate}
\renewcommand{\arraystretch}{1.1}
\begin{tabular}{l|l|l|l}
\hline\hline
          & \mc{1}{c|}{\tb{Anchor 1}}
          & \mc{1}{c|}{\tb{Anchor 2}}
          & \mc{1}{c}{$\bias^r$}\\ \hline
\tb{True values}     & 15.000, \ 0.000, 1.250  & 7.500, -5.000, 1.500 &0.050 \\ \hline
\tb{Initial values}  & 15.020, \ 0.000, 1.000  & 7.450, -5.070, 1.000 &0.000 \\ \hline
{nbh\_01 est.}       & 15.018, \ 0.019, 1.245  & 7.449, -5.046, 1.510 & 0.026 \\
{nbh\_02 est.}       & 15.018, \ 0.038, 1.243  & 7.456, -5.038, 1.484 & 0.021 \\
{nbh\_03 est.}       & 15.011, \ 0.109, 1.506  & 7.467, -4.995, 1.706 & 0.008 \\
{nbh\_04 est.}       & 15.018, \ 0.025, 1.253  & 7.451, -5.043, 1.502 & 0.031 \\
{nbh\_05 est.}       & 15.018, \ 0.055, 1.253  & 7.462, -5.032, 1.449 & 0.020 \\
\hline\hline
\end{tabular}
\end{table}

\begin{figure}
    \centering
    \includegraphics[width=\linewidth]{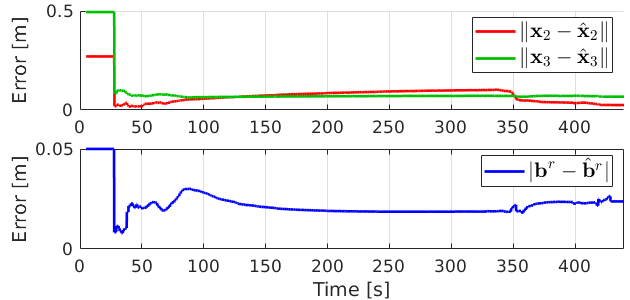}
    \caption{Error of the estimates on anchor position and ranging bias by the BA process over time.}
    \label{fig: uwb exin est}
\end{figure}

From \tref{tab: all ATE} we can clearly see that VIRAL SLAM consistently achieves better performance compared to existing methods in all datasets, even when only one lidar is used. For VINS-Mono, VINS-Fusion and VIRAL SLAM, we also report the BA results, i.e. the positions of the key frames refined by loop closure and BA process. We note that the pure odometry and BA results of VIRAL SLAM do not differ much, while there is a large difference between odometry and BA results of the VIO methods. This can be explained as VIRAL SLAM has much less drift than VINS methods, thus there is not a lot of correction made in the BA process. \fref{fig: viral slam eee 02} presents the result of VIRAL SLAM in one experiment. Also, we can see from \tref{tab: anchor estimate} and \fref{fig: uwb exin est} that the BA process does bring down the error in the extrinsic and intrinsic parameters of the UWB network.

\subsection{Ablation study}

We further conduct extra experiments with different subsets of sensor suites to study the contribution of each sensor in the overall localization scheme. Tab. \ref{tab: all ATE ablation} reports the ATE of the key frame positions of VIRAL SLAM over the experimented datasets. From this table we can see that the lidar-only and lidar-camera setups has roughly similar average ATE over the datasets. From lidar-only to lidar-range, there is a significant increase in error. This can be explained as that with UWB we are not just estimating the key frame poses, but also states relating to the UWB anchor position and bias. Thus the increase in ATE is a trade-off of knowledge on the UWB states. Interestingly, when comparing the ATE of lidar-UWB with lidar-camera-UWB setup, we can observe a decrease in ATE. Thus, we can conclude that the loop closure constraints introduced by using camera do help bring down the error in the BA process.

\section{Conclusion and Future Works} \label{sec: conclusion}

In this paper we have developed a multi-sensor SLAM method leveraging an extensive set of sensors: stereo camera, lidar, IMU, UWB, so-called VIRAL SLAM. The system features synchronization and integration of multiple lidars with complementary field of view (FOV), depth-matching of stereo camera visual features with pointcloud map, vision-triggered lidar-refined loop closure, UWB extrinsic and intrinsic parameter estimation. Via extensive experiments results, we have demonstrated that VIRAL SLAM can achieve highly accurate localization results as well as robustness in challenging conditions.

\balance
\bibliographystyle{IEEEtran}
\bibliography{references}

\end{document}